\documentclass{article} 
\usepackage{iclr2027_conference,times}
\usepackage{multirow}
\usepackage{multicol}
\usepackage{xcolor}
\usepackage{booktabs}
\usepackage{amsmath,amsfonts}
\usepackage{algorithmic}
\usepackage{algorithm}
\usepackage{array}
\usepackage{float}
\usepackage{booktabs}
\usepackage{subcaption}
\usepackage[caption=false,font=normalsize,labelfont=sf,textfont=sf]{subfig}
\usepackage{textcomp}
\usepackage{stfloats}
\usepackage{url}
\usepackage{verbatim}
\usepackage{graphicx}
\usepackage{cite}

\usepackage{amsmath,amsfonts,bm}

\def\eqref#1{equation~\ref{#1}}

\def\1{\bm{1}}

\DeclareMathAlphabet{\mathsfit}{\encodingdefault}{\sfdefault}{m}{sl}
\SetMathAlphabet{\mathsfit}{bold}{\encodingdefault}{\sfdefault}{bx}{n}

\title{PAVXploreRL: Physical-Action-Visual World Model Reinforcement Learning with Action Exploration}

\author{
\shortstack[l]{
Han Wang$^{1}$,
Zijun Wang$^{2}$,
Shuoshuo Xue$^{2}$,
Rui Cao$^{1}$,
Fengjiao Chen,\\
Xiaodan Liang$^{2}$,
Roy Ka-Wei Lee$^{1}$
}\\[1em]
$^{1}$Singapore University of Technology and Design\\
$^{2}$Sun Yat-sen University
}
\iclrfinalcopy 
\begin{document}

\maketitle
\begin{abstract}
Action-conditioned world models are a key component of embodied AI, serving as scalable policy evaluators that reduce reliance on expensive real-world rollouts. To accurately capture diverse action-induced dynamics, such models should satisfy three key objectives—Physical Plausibility (P), Action Adherence (A), and Visual Fidelity (V), collectively referred to as PAV—while remaining robust to both in-distribution (ID) expert demonstrations and out-of-distribution (OOD) actions. 
However, existing methods primarily rely on ID action–video pairs and pixel-level reconstruction losses, which do not explicitly optimize PAV objectives and generalize poorly beyond expert data.
To address this, we propose \textbf{PAVXploreRL}, a reinforcement learning framework built on a pretrained latent world model that explicitly optimizes PAV objectives through reward-driven training. To improve action generalization, our method jointly leverages ID trajectories and noise-driven OOD action exploration, without paired video supervision. 
Experiments show that PAVXploreRL consistently outperforms pretrained baselines, achieving a 5.6\% average gain across benchmarks and producing higher-quality PAV properties. As a policy evaluator, it also yields more reliable performance estimates and reduces the overestimation bias of prior expert-only world models such as Ctrl-World. 
Code: \url{https://github.com/Social-AI-Studio/PAVXploreRL}

\end{abstract}

\section{Introduction}

With the advent of large-scale video generation models, controllable world models have emerged as a promising paradigm for embodied AI, predicting future observations from initial states and control signals to better emulate real-world dynamics. \citep{guo2025ctrlworld,gao2026dreamdojo,chen2026abotphysworld,liao2025genie,zhu2024irasim,quevedo2025evaluating}. 
They can be broadly categorized into text-conditioned models, which function as action policies, and action-conditioned models, which directly take robot actions as input to simulate future observations under real-world dynamics. The latter are commonly used as virtual environments for policy evaluation and development, reducing the cost and risk of real-world deployment. This work focuses on action-conditioned world models.

Action-conditioned world models are expected to satisfy the Physical-Action-Visual (PAV) targets \citep{reworld2026,wang2026world}. First, \textbf{Physical Plausibility}: they must generate physically plausible trajectories that respect real-world dynamics, including object permanence, consistent robot–object contacts, and valid rigid or deformable body behavior, as violations lead to unrealistic simulations. Second, \textbf{Action Adherence}: models should faithfully follow conditioning actions, as weak controllability undermines their reliability as policy evaluators. Third, \textbf{Visual Fidelity}: outputs should maintain high visual fidelity, preserving sufficient detail for human interpretation and downstream tasks. Most existing approaches, however, rely primarily on pixel-level similarity targets or non-adapted general VLM scores, providing limited explicit supervision for PAV targets.

Another limitation of current world models is their reliance on in-distribution (ID) action–video pairs collected from human experts, which predominantly consist of successful task-completion trajectories\citep{wu2026rlvr,bardhan2026persistent,mindv2026}. As a result, current world models may overestimate the effectiveness of out-of-distribution (OOD) actions from suboptimal policies, inflating policy success rates. 
Reliable world models must generalize across diverse policies and accurately predict future outcomes under unseen actions; however, the absence of corresponding videos for such OOD actions limits their integration into standard training pipelines. 

\begin{figure}[!t]
\centering
\small
\includegraphics[width=0.90\columnwidth]{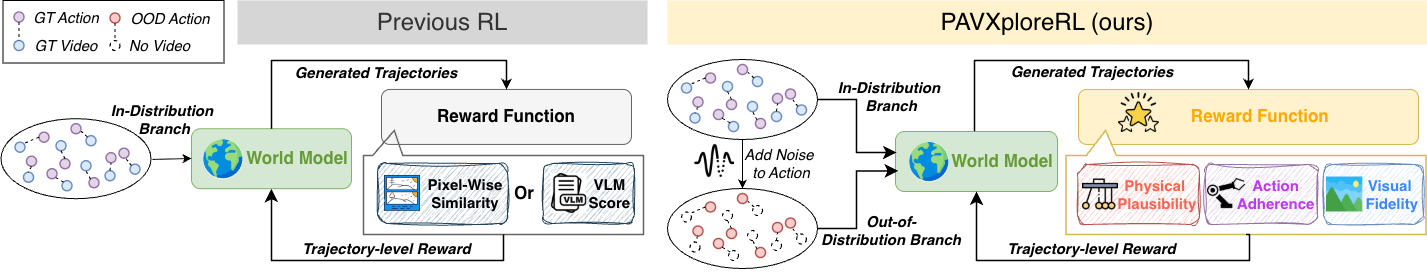}
\caption{Comparison between \textbf{PAVXploreRL} and prior RL: unlike in-distribution-only expert-based approaches, it enables both in- and out-of-distribution action exploration and replaces pixel/VLM rewards with a domain-adapted latent world model for direct PAV objective evaluation.}
\label{fig:rl_overview}
\end{figure}

To address these challenges, we propose \textbf{PAVXploreRL} (Fig.~\ref{fig:rl_overview}, right), a reinforcement-learning method that enhances the PAV properties of action-conditioned world models via action exploration. Built upon the latent world model VJEPA-2~\citep{mido2025vjepa2}, which demonstrates strong physical understanding compared to other latent or vision-language models, we design three complementary rewards targeting a distinct PAV component: (1) a raw VJEPA reward for physical consistency, (2) an embodiment-focused VJEPA reward for accurate action execution, and (3) a visual fidelity reward perceptually realistic frames. 
Leveraging VJEPA-2’s latent predictive capabilities, these rewards can be computed for both ID action–video pairs and OOD action generated via perturbations of ID actions with control noise, where paired videos are unavailable.
To enable the above function, we retrain the action-conditioned VJEPA-2 on our robotic dataset using both raw and embodiment-focused frames, enabling multi-step autoregressive prediction over the whole action chunk.

To validate our approach, we build an action-conditioned world model by augmenting a state-of-the-art video diffusion model with action encoding and temporal history modeling. After supervised fine-tuning (SFT) on robotic datasets to obtain the pretrained baseline, we further post-train the model using our PAVXploreRL framework. Experiments yield consistent gains over the pretrained baseline, averaging 5.6\% across all benchmarks. Qualitative results further indicate improved physical plausibility, action adherence, and visual realism. Experiments as policy evaluator also show reduced policy overestimation after our OOD action exploration.
We summarize our contributions:
\textbf{(1).} We propose a unified reward function that explicitly optimize the PAV targets-Physical Plausibility, Action Adherence, and Visual Fidelity.
\textbf{(2).} We develop an RL framework that leverages both ID action–video pairs and OOD actions without paired video supervision.
\textbf{(3).} We show that PAVXploreRL consistently improves pretrained world models, yielding more reliable policy evaluation.

\section{Related Works}
\textbf{World Model:}
Recently, advances in large-scale video diffusion models~\citep{agarwal2025cosmos,agarwal2026cosmos,wan2025open,chi2025wow} have made generated videos increasingly realistic and controllable. Building on these, robotic world models have also progressed rapidly. 
According to the control signal, existing methods can be categorized into two groups. The first is text-conditioned models, which either implicitly~\citep{jang2025dreamgen,bharadhwaj2024gen2act,team2025gigaworld0} or explicitly~\citep{du2024universal,hu2024videoprediction,liang2024dreamitate,liao2025genie,tan2025anypos,feng2025vidar,Ye2025GigaBrain0AW,chen2026abotphysworld} function as action policy. Some recent works further unify policy learning and world modeling through co-training~\citep{zhao2025cotvla,li2025unifiedvideoaction,zhu2025unifiedworld,guo2024prediction,gao2024flip,zhang2025upvla,zheng2025flare,zhong2025flowvla,ye2026world,bi2026motus}. 
The second is action-conditioned world models, which take robot actions as input and predict future observations consistent with real world interactions~\citep{zhu2024irasim,bardhan2026persistent,guo2025ctrlworld,gao2026dreamdojo,zhang2025reinforcing,team2026advancing}. These models can serve as policy evaluators. Moreover, integrating action-conditioned world models into policy training reduces reliance on real-robot rollouts and improves task success rates.~\citep{guo2025ctrlworld,zhang2025reinforcing,zhang2026practicalwrl,jiang2026wovr,sharma2026worldgymnast}.

\textbf{Physical-Action-Visual Targets Evaluation:}
Previous works have identified key criteria for world models, including physical realism, task completion, embodiment plausibility, and visual quality~\citep{reworld2026,wang2026world}.
 Building on this, we propose the Physical-Action-Visual (PAV) targets for action-conditioned world models.
\textbf{Physical Plausibility} is often assessed by pretrained VLMs~\citep{reworld2026,chen2026abotphysworld}, but their physical understanding is limited without domain-specific training. Alternatively, VJEPA—a latent world model—has demonstrated superior physical understanding compared to VLM-based methods~\citep{bardes2024featureprediction,mido2025vjepa2,garrido2025physics} and improves video generation quality during both inference~\citep{yuan2026inference} and training~\citep{mindv2026}. To align with our task, we train an action-conditioned VJEPA-2~\citep{mido2025vjepa2} on our robot data and use it to measure the physical plausibility.
\textbf{Action adherence} is commonly evaluated using inverse dynamics models (IDMs)~\citep{tian2025predictive,chi2025wow}. However, IDMs infer actions from frame differences, inducing an inverse information flow relative to forward world models. To maintain consistency, we use the embodiment segmentation strategy of~\citep{chi2025wow} to adapt the action-conditioned VJEPA-2 to explicitly measure alignment between robot motion and actions.
\textbf{Visual fidelity} is commonly assessed via pixel-wise reconstruction loss. As our out-of-distribution (OOD) branch lacks paired videos, we instead design a reference-free reward integrating perceptual, statistical, and temporal metrics, representative metrics such as NIQE~\citep{mittal2013making}, SSIM~\citep{wang2004image}, and temporal consistency~\citep{lai2018learning}.

\textbf{Reinforcement Learning of World Models.} 
Prior RL approaches for world models, such as Flow-GRPO~\citep{liu2026flowgrpo}, rely on log-likelihood estimation and storing full denoising trajectories, resulting in high memory usage and prolonged training. More efficient methods—FPO~\citep{mcallister2025flow} and Diffusion-NFT~\citep{zheng2025diffusionnft}—avoid log-likelihood computation and have been integrated into recent world model training pipelines~\citep{wang2026worldcompass,bardhan2026persistent}. So our paper would adopt the Diffusion-NFT~\citep{zheng2025diffusionnft}. However, previous action-conditioned world model RL still depends on ground-truth (GT) video for reward evaluation \citep{wu2026rlvr,bardhan2026persistent,mindv2026}, limiting exploration of OOD actions. We overcome this by introducing a GT-free reward based on latent future predictions capability of our main reward evaluator, VJEPA-2, enabling RL with real action exploration including OOD actions.

\section{Method}
\begin{figure*}[!t]
\centering
\includegraphics[width=0.85\textwidth]{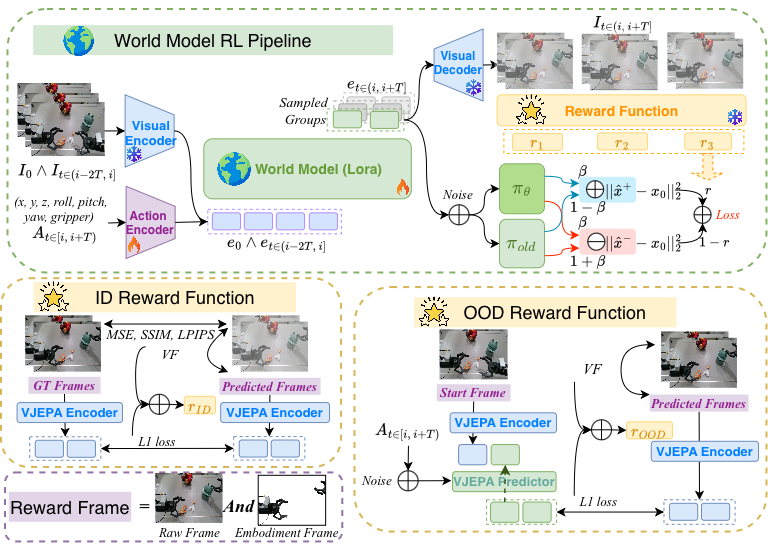}
\caption{Overview of the \textbf{PAVXploreRL} framework. \textbf{World Model Reinforcement Learning Pipeline:} Given historical frames and actions, predicted future frames are evaluated by the reward function, and the resulting rewards weight the diffusion policy optimization loss. \textbf{In-distribution Reward Function:} For in-distribution ground-truth (GT) action–video pairs, rewards combine pixel-level metrics (MSE, SSIM, and LPIPS), VJEPA encoder-based metrics, and the proposed visual fidelity (VF) metrics. \textbf{Out-of-distribution Reward Function:} For out-of-distribution actions without GT videos, rewards are computed using VJEPA predictor-based metrics and VF metrics. \textbf{Reward Frame:} VJEPA-based rewards are assessed on both raw and embodiment-focused frames.} 
\label{fig:rl_pipeline}
\end{figure*}


An overview of our framework is shown in Figure~\ref{fig:rl_pipeline}.  We discuss our model architecture and training in Section~\ref{sec:wm_design}, the reward function design in Section~\ref{sec:rw_function_design}, and the VJEPA-2 training in Section~\ref{sec:vjepa_training}.

\subsection{World Model Architecture and Training}
\label{sec:wm_design}

We define the world model $\pi$ as a conditional video predictor. At step $t=i$, given the first frame $I_0$ of video, history frames $I_{t \in (i-2T, i]}$, and future actions $A_{t \in [i, i+T)}$, the model predicts future frames $\hat{I}_{t \in (i, i+T]}$. The policy is trained to minimize the discrepancy between predicted and real frames:
\begin{equation}
\pi\big(I_0, I_{t \in (i-2T, i]}, A_{t \in [i, i+T)}\big) = \hat{I}_{t \in (i, i+T]} \approx I_{t \in (i, i+T]}
\end{equation}
We build our action-conditioned world model by adapting a pretrained video diffusion model with two modifications: incorporating action conditioning and maintaining a compact history representation for efficient long-horizon interaction.
First, we introduce action conditioning by injecting action embeddings into the temporal pathway via adaptive layer normalization (AdaLN) and time embeddings, enabling action controllability. 
Second, to capture long-range temporal context, we adopt a Framepack~\citep{zhang2026framecontext} strategy that samples and compresses history frames in $0 \land (i-2T, i]$ unevenly—distant frames are sampled densely and heavily compressed. 
The model is first trained with supervised fine-tuning (SFT) using a reconstruction MSE loss; upon convergence (no further gains on the validation set with additional training), it is further optimized via reinforcement learning (RL) by updating only the action encoder and LoRA parameters.

Our RL training pipeline is illustrated in the top of Figure~\ref{fig:rl_pipeline}. We adopt the DiffusionNFT~\citep{zheng2025diffusionnft} framework to efficiently fine-tune diffusion-based world models. 
During rollouts, we generate $G$ groups of future frames using a first-order stochastic differential equation (SDE) solver conditioned on historical frames and future actions. The generated frames  are evaluated with our reward function $R$, yielding a scalar trajectory-level reward $r \in [0,1]$. For training, we maintain two model snapshots: (i) current policy $\pi_{\theta}$ and (ii) EMA-updated policy $\pi_{\text{old}}$.
 Implicit positive and negative denoising targets are constructed as:
\begin{equation}
\hat{\mathbf{x}}_0^{+} = (1-\beta)\hat{\mathbf{x}}_0^{\text{old}} + \beta \hat{\mathbf{x}}_{0,\theta}, 
\qquad
\hat{\mathbf{x}}_0^{-} = (1+\beta)\hat{\mathbf{x}}_0^{\text{old}} - \beta \hat{\mathbf{x}}_{0,\theta}
\end{equation}
where $\hat{\mathbf{x}}_0^{\text{old}} \sim \pi_{\text{old}}$ and $\hat{\mathbf{x}}_{0,\theta} \sim \pi_{\theta}$. Positive targets encourage reward-aligned paths, while negative targets suppress deviations toward suboptimal paths. The training objective is a reward-weighted denoising loss:
\begin{equation}
\begin{split}
\mathcal{L}(\theta) &= 
\mathbb{E}\Big[ r \, \|\hat{\mathbf{x}}_0^{+} - \mathbf{x}_0\|_2^2 + (1-r)\, \|\hat{\mathbf{x}}_0^{-} - \mathbf{x}_0\|_2^2 \Big] 
\end{split}
\end{equation}
where $x_0 = \hat{I}_{t \in (i, i+T]}$ is the predicted frames in rollout stage.

\subsection{Reward Function Design}
\label{sec:rw_function_design}

We define a unified reward function $\mathcal{R}$ that maps a predicted frame sequence to a scalar score $r \in [0,1]$, reflecting video generation quality. As illustrated in the bottom of Figure~\ref{fig:rl_pipeline}, the reward formulation depends on supervision availability, combining an in-distribution (ID) branch $r_{\text{ID}}$ with ground-truth action–video pairs and an out-of-distribution (OOD) branch $r_{\text{OOD}}$ with only perturbed actions without video pairs. The branch is selected via a Bernoulli-like sampling scheme:
\begin{equation}
r =
\begin{cases}
r_{\text{OOD}}, & \text{if } \epsilon \le \alpha_{\text{OOD}},\\
r_{\text{ID}}, & \text{otherwise},
\end{cases}
\quad \epsilon \sim \mathrm{Uniform}(0,1).
\end{equation}
where $\alpha_{\text{OOD}}$ determines the proportion of OOD trajectories.
To generate suboptimal actions without incurring the computational cost of policy inference during training, we construct OOD actions by applying three perturbations to non-static GT action segments exceeding a motion threshold, treating each hand and action type (position, rotation, and gripper) independently: \textbf{(1) Temporal shift:} segment boundaries $(s,e)$ are randomly perturbed within $\pm 0.6$ s;  \textbf{(2) Structured magnitude bias:} chunk-wise multiplicative scaling induces consistent amplitude deviations of $\pm 0.15$; and \textbf{(3) Smooth temporal noise:} Gaussian noise ($\sigma=0.02$) is added at each timestep and temporally accumulated to produce smooth drift. Human inspection indicates that the resulting trajectories deviate substantially from GT actions.

The unified reward comprises five components, each targeting a different aspect of video quality:

\textbf{Pixel-level Reward} $r_{\text{pixel}}$:
This reward measures image similarity between predicted and GT frames using MSE, SSIM, and LPIPS, providing low-level supervision via direct comparison. After group-level normalization, the three metrics are averaged to obtain the final reward:
\begin{equation}
r_{\text{pixel}} =
\frac{1}{3}
\left(
r_{\text{MSE}} +
r_{\text{SSIM}}+
r_{\text{LPIPS}}
\right)
\end{equation}
Normalization formula to $[0,1]$ is given as follows and applied to all forward reward normalization.
\begin{equation}
r_i^{\mathrm{norm}}
=
0.5
+
0.5 \cdot
\mathrm{clip}
\!\left(
\frac{r_i-\mu_G}{\sigma_G+\epsilon},
-1,1
\right),\quad
\mu_G = \mathrm{mean}_{j \in G}(r_j), \qquad
\sigma_G = \mathrm{std}_{j \in G}(r_j).
\end{equation}
\textbf{VJEPA Encoder Reward} $r_{\text{VJEPA-E}}$:
We leverage VJEPA-2's strong physical understanding capability to supervise high-level properties. For GT action-video pairs, we compute an encoder-based reward using the VJEPA encoder $E(\cdot)$ to measure the similarity between predicted and GT frames:
\begin{equation}
r_{\text{VJEPA-E}}
=
-
\left\|
E(\hat{I}_{t \in (i, i+T]}) - 
E(I_{t \in (i, i+T]}) 
\right\|_1
\end{equation}
\textbf{VJEPA Predictor Reward} $r_{\text{VJEPA-P}}$:
To evaluate physical plausibility without GT frames, we use an action-conditioned VJEPA-2's predictor $P(\cdot)$, which predicts future latent representations from the start frame $I_i$ and actions $A = A_{t \in [i, i+T)}$. We compute the predictor-based reward by comparing the world model predicted frames with the VJEPA-based latent prediction:
\begin{equation}
r_{\text{VJEPA-P}}(A)
=
-
\left\|
E(\hat{I}_{t \in (i, i+T]}) -
P\big(E(I_i), A_{t \in [i, i+T)}\big)
\right\|_1
\end{equation}
However, latent prediction reward tends to favor static but high-quality predictions. To mitigate this, we introduce a static-motion regularization term using a zero-action sequence $A_0$. Specifically, the latent representation predicted under $A_0$ should remain static and dissimilar to the world model generated videos. The final predictor-based reward is given by:
\begin{equation}
r_{\text{VJEPA-P}}
=
r_{\text{VJEPA-P}}(A)
-
\lambda_{\text{static}} r_{\text{VJEPA-P}}(A_0)
\end{equation}
where $\lambda_{\text{static}} \in [0, 1]$ controls the strength of the static regularization.

\textbf{Embodiment-focused VJEPA Reward} $r_{\text{VJEPA-E/P}}^{\text{emb}}$:
To better assess action adherence, both encoder- and predictor-based rewards are computed on two complementary visual streams: (1) raw frames and (2) embodiment-focused frames. While raw-frame rewards may be influenced by background dynamics and object interactions, embodiment-focused rewards emphasize robot motion consistency with executed actions. The final rewards are combinations of both streams:
\begin{equation}
r_{\text{VJEPA-E/P}}
=
r_{\text{VJEPA-E/P}}^{\text{raw}}
+
\lambda_{\text{emb}}
r_{\text{VJEPA-E/P}}^{\text{emb}}
\end{equation}
where $\lambda_{\text{emb}} \in [0, 1]$ controls the contribution of the embodiment-focused reward.

\textbf{Visual Fidelity Reward} $r_{\text{VF}}$:
As a latent model, VJEPA-2 does not explicitly preserve fine image details. To complement this, we introduce $r_{\text{VF}}$, which combines frame-level and temporal no-reference metrics to capture sharpness, perceptual quality, motion consistency, structural coherence, and color stability. (Details of each component is in Appendix~\ref{sec:visual_quality_reward}). After normalization, all components are averaged to obtain the visual fidelity reward.

\textbf{Final Reward Combinations}
After group-level normalization, the final reward is defined as:
\begin{equation}
r_{\text{ID}} = \lambda_{\text{pixel}}\, r_{\text{pixel}} + \lambda_{\text{VJEPA}}\, r_{\text{VJEPA-E}} + \lambda_{\text{VF}}\, r_{\text{VF}}, \quad
r_{\text{OOD}} = (1-\lambda_{\text{VF}})\, r_{\text{VJEPA-P}} + \lambda_{\text{VF}}\, r_{\text{VF}}
\end{equation}
where $\lambda_{\text{VJEPA}}$ and $\lambda_{\text{VF}}$ control the VJEPA and VF contributions, and $\lambda_{\text{pixel}} = 1 - \lambda_{\text{VJEPA}} - \lambda_{\text{VF}}$.

\subsection{VJEPA-2 Training Pipeline}
\label{sec:vjepa_training}

To adapt VJEPA-2 to the robotic domain, we fine-tune its action-conditioned variant on robotic data with two modifications. First, while standard VJEPA-2 captures holistic scene-level representations, embodied settings require explicit modeling of the agent’s structure (e.g., grippers) for accurate dynamics. We therefore introduce two complementary inputs: (i) raw images and (ii) embodiment-focused images obtained via segmentation, retaining only robot-specific regions.
Second, VJEPA-2's original one-step prediction limits temporal consistency and robustness over longer horizons. While accumulative training exists in the original implementation, it is still insufficient for long-horizon prediction. We therefore adopt an accumulative loss over multi-step predictions across the entire predicted frame chunk $T$, enabling the model to predict future frames given only the start frame and corresponding actions, akin to standard world models.

To preserve the VJEPA encoder’s physical understanding, we adopt its action-conditioned training setup, freezing the encoder and training an action-conditioned predictor from scratch.
And to accelerate training, we split the training into two stages, the pipeline is as shown in Appendix~\ref{sec:vjepa_training_appendix}.
In Stage 1, following standard action-conditioned VJEPA-2 training, the model performs one-step future prediction, using frames $I_i,\ldots,I_{i+T-1}$ and actions to predict only $I_{i+T}$ embedding. In Stage 2, we introduce two modifications: (i) random sampling from raw and embodiment-focused frames, and (ii) autoregressive rollout from start frame $I_i$ and $A_{t \in [i, i+T)}$ with accumulated loss over horizon $T$, improving long-horizon consistency and embodiment sensitivity.

\section{Experiments}
\begin{table*}[!t]
\centering
\small
\setlength{\tabcolsep}{4pt}
\renewcommand{\arraystretch}{1.2}
\caption{Quantitative comparison of interactive long-horizon trajectory generation on the Agibot and Droid validation set. Our models generate 9.6-second trajectories autoregressively from a start frame conditioned on three sequential 3-step action chunks. Other baseline models follow their respective training settings.
$^{*}$: Reproduced results. PT: Pretrained; RL: Reinforcement Learning; CS: Control Signal; T: Text; AC: Action.}
\begin{tabular}{c|l|c|ccc|cc|cc}
\hline
\multirow{2}{*}{Dataset} & \multirow{2}{*}{Method}  & \multirow{2}{*}{CS} 
& \multicolumn{3}{c|}{Visual Quality} 
& \multicolumn{2}{c|}{Flow Consistency} 
& \multicolumn{2}{c}{VJEPA Loss} \\
\cline{4-10}
& 
& &   PSNR $\uparrow$ & SSIM $\uparrow$ & LPIPS $\downarrow$
& EPE $\downarrow$ & COS $\uparrow$
& Enc $\downarrow$ & Pred $\downarrow$ \\
\hline

\multirow{5}{*}{Agibot}
& Genie-Envisioner$^{*}$                & AC  & 17.44 & 0.730 & 0.159 & 0.514 & 0.021 & 0.545 & - \\
& ABot-PhysWorld$^{*}$                 & T  & 18.24 & 0.772 & 0.189 & 0.516 & 0.073 & 0.526 & - \\
& DreamDojo$^{*}$              & AC  & 19.69 & 0.790  & 0.161 & 0.437 & 0.155 & 0.528 & - \\

& PT (ours)              & AC  & 20.52 & 0.793 & 0.133 & 0.346 & 0.203 & 0.524 & 0.379 \\
& RL (ours)     & AC & \textbf{21.20} & \textbf{0.812} & \textbf{0.127} & \textbf{0.306} & \textbf{0.231} & \textbf{0.521} & \textbf{0.373} \\
\hline

\multirow{6}{*}{Droid}
& WPE                & AC  & 20.33 & 0.772 & 0.131  & - & - & - & -  \\
& IRASIM                &  AC & 21.36 & 0.774 & 0.117  & - & - & - & -  \\
& Ctrl-World               &  AC  & 21.27 & 0.793 & 0.110 & - & - & - & - \\
& Ctrl-World$^{*}$              &  AC  & 21.30 & 0.827 & 0.097 &  \textbf{0.222} & 0.162 & \textbf{0.509} & - \\
& PT (ours)             & AC  & 20.79 & 0.814 & 0.097 & 0.252 & 0.168 & 0.511 & 0.373 \\
& RL (ours)    & AC  & \textbf{21.71} & \textbf{0.828} & \textbf{0.092} & 0.227 & \textbf{0.197} & \textbf{0.509} & \textbf{0.369} \\
\hline
\end{tabular}

\label{tab:main_results}
\end{table*}

\begin{figure*}[!t]
\centering
\includegraphics[width=0.85\textwidth]{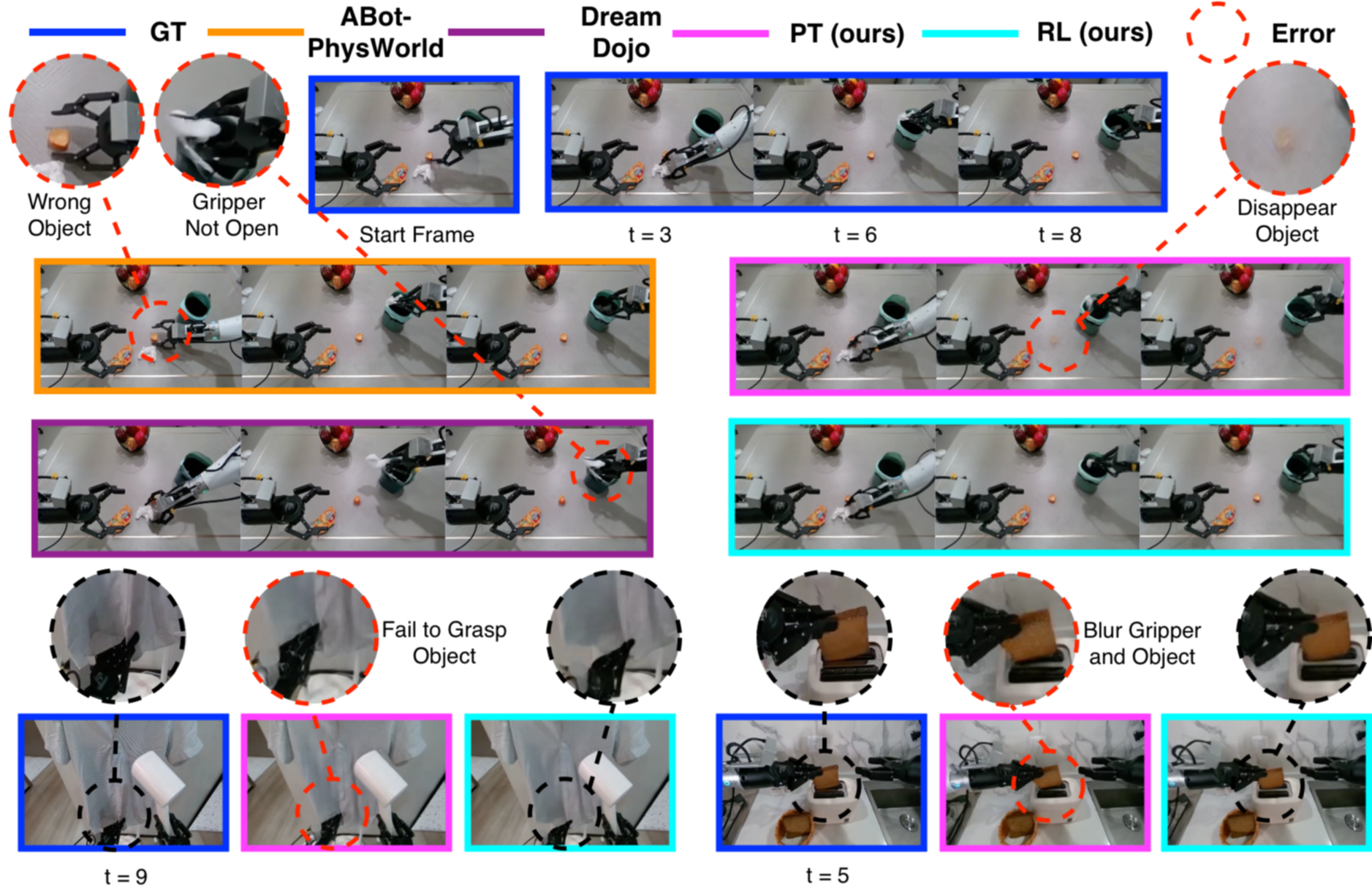}
\caption{Qualitative comparison of long-horizon rollouts on Agibot validation set.
} 
\label{fig:demo_example}
\end{figure*}

\begin{table*}[t]
\centering
\small
\setlength{\tabcolsep}{4pt}
\renewcommand{\arraystretch}{1.15}
\caption{
Reward-function ablations on the AgiBot validation set after 1K RL
steps. We separately evaluate the ID and OOD reward branches and
remove individual auxiliary components from the full PAV reward.
}
\label{tab:reward_ablation}

\begin{tabular}{lcc|ccc|cc|cc}
\toprule
\multirow{2}{*}{\textbf{Setting}}
& \multicolumn{2}{c|}{\textbf{Reward Branch}}
& \multicolumn{3}{c|}{\textbf{Visual Quality}}
& \multicolumn{2}{c|}{\textbf{Flow Consistency}}
& \multicolumn{2}{c}{\textbf{VJEPA Loss}} \\
\cline{2-10}

& ID & OOD
& PSNR $\uparrow$ & SSIM $\uparrow$ & LPIPS $\downarrow$
& EPE $\downarrow$ & COS $\uparrow$
& Enc $\downarrow$ & Pred $\downarrow$ \\
\midrule

Pixel only
& $\times$ & $\times$
& 20.80 & 0.806 & 0.145
& 0.333 & 0.204
& 0.526 & 0.374 \\

Pixel + OOD
& $\times$ & $\checkmark$
& 20.84 & 0.805 & 0.131
& 0.325 & 0.227
& 0.523 & 0.375 \\

Pixel + ID
& $\checkmark$ & $\times$
& 20.93 & 0.807 & 0.135
& 0.324 & 0.216
& 0.524 & 0.374 \\

\midrule

Full w/o Static
& $\checkmark$ & $\checkmark$
& 20.85 & 0.806 & 0.140
& 0.328 & 0.212
& 0.526 & 0.409 \\

Full w/o Emb
& $\checkmark$ & $\checkmark$
& 20.89 & 0.807 & 0.139
& 0.326 & 0.213
& 0.525 & 0.374 \\

Full w/o VF
& $\checkmark$ & $\checkmark$
& 20.94 & 0.807 & 0.128
& 0.318 & 0.218
& 0.525 & 0.374 \\

\midrule

\textbf{Full PAV reward}
& $\checkmark$ & $\checkmark$
& \textbf{21.05} & \textbf{0.810} & \textbf{0.126}
& \textbf{0.310} & \textbf{0.230}
& \textbf{0.521} & \textbf{0.373} \\

\bottomrule
\end{tabular}
\end{table*}

\textbf{Datasets:} Our primary dataset is Agibot Alpha \citep{bu2025agibot}, a dual-hand manipulation dataset with 92K trajectories (~590 h) across 36 tasks. To evaluate dataset robustness, we also use DROID dataset \citep{khazatsky2024droid}, a single-hand dataset with 76K trajectories (~350 h) over 86 tasks.

\textbf{Training Details:} 
We adopt Wan2.2-TI2V-5B~\citep{wan2025open} as the base model, conditioned on the first frame and 16 sparse history frames (6.4 s) to predict future $T = 16$ frames at 5 FPS (3.2 s). Input resolutions are $288\times384$ (Agibot) and $192\times320$ (Droid). The model uses single-view (Agibot: head view; Droid: third-person view) observations and 14D action inputs, consisting of 6D end-effector poses (x, y, z, roll, pitch, yaw) and gripper states per hand, with single-hand action zero-padded.
The unified reward function uses $\alpha_{\text{OOD}} = 0.1$, $\lambda_{\text{VJEPA}} = 0.2$, $\lambda_{\text{static}} = 0.8$, $\lambda_{\text{emb}} = 0.5$, and $\lambda_{\text{VF}} = 0.1$, selected based on the ablation study in Section~\ref{sec:ablation_study}.
Full supervised fine-tuning (SFT) is conducted on 8×H200 GPUs for 80K steps (batch size 32, ~4 epochs, ~3 days). Reinforcement learning (RL) further updates the action encoder and LoRA modules on the same hardware for 5K steps (group size 12, batch size 8, ~5 days). Validation performance over training steps is shown in Appendix~\ref{sec:model_performance}.
Stage 1 VJEPA-2~\citep{mido2025vjepa2} training requires 60K steps (batch size 64, ~3 days), while Stage 2 requires 12K steps (batch size 64, ~5 days). 
To extract the embodiment-only frame, we uses GroundingDINO~\citep{liu2024grounding} for robot detection and Segment Anything Model (SAM)~\citep{kirillov2023segment} for robot segmentation.

\textbf{Baselines and Evaluation Metrics:} 
For Agibot dataset, we evaluate single-view models with published weights, having Genie-Envisioner~\citep{liao2025genie}, ABot-PhysWorld~\citep{chen2026abotphysworld}, and DreamDojo~\citep{gao2026dreamdojo}; for methods without reported results, we re-run evaluations. We hold out 1\% per task for validation and sample 3 trajectories per task to test, yielding ~220 video clips of 9.6 s each.
For DROID dataset, we follow the Ctrl-World validation set and report results from prior single-view methods, including WPE~\citep{quevedo2025evaluating}, IRASIM~\citep{zhu2024irasim}, and Ctrl-World~\citep{guo2025ctrlworld}, with Ctrl-World reproduced.
Evaluation metrics include visual quality metrics (PSNR \citep{hore2010psnrssim}, SSIM \citep{wang2004image}, LPIPS \citep{zhang2018unreasonable}) and flow-based metrics (Avg EPE and COS \citep{zhang2025reinforcing}), where EPE measures motion magnitude error and COS measures directional consistency between predicted and ground-truth frames. We further report VJEPA-2-based losses, including encoder loss under GT actions and predictor loss under the same suboptimal actions.

\subsection{Main Results}
As shown in Table~\ref{tab:main_results}, our RL pipeline consistently improves upon the pretrained (PT) model, particularly in SSIM, LPIPS, and flow consistency, indicating enhanced structural fidelity, motion alignment and action controllability. Across both Agibot and DROID, our method achieves the best overall performance among all single-view baselines. Notably, although the PT model trails several baselines on DROID, RL training enables it to surpass them, highlighting the robustness and generalizability of our pipeline across datasets.

We further present qualitative comparisons in Figure~\ref{fig:demo_example} on the Agibot dataset using four representative models. As shown in the top example, despite its strong visual realism enabled by a 14B-parameter scale, ABot-PhysWorld~\citep{chen2026abotphysworld} frequently fails to follow the text instructions, such as grasping incorrect objects. DreamDojo~\citep{gao2026dreamdojo} exhibits weak long-horizon consistency, where action execution deteriorates during autoregressive rollouts, leading to delayed or missing gripper motions. Our pretrained (PT) model demonstrates reasonable action adherence but still suffers from limitations. In the top example, it violates physical plausibility—object permanence—having objects to disappear during gripper-induced occlusions. In the bottom-left example, it violates action adherence, as fine-grained manipulation actions, such as grasping a specific corner of a cloth, are occasionally missed. In the bottom-right example, visual fidelity degrades due to motion blur during object interaction. In contrast, our RL-enhanced model consistently achieves superior physical plausibility, action adherence, and visual fidelity across all examples, demonstrating the effectiveness of the proposed unified reward design.

A blinded human evaluation on 100 randomly sampled AgiBot trajectories
further confirms that PAVXploreRL is preferred over the pretrained
model in physical plausibility, action adherence, visual fidelity, and
overall quality; detailed protocols and results are provided in
Appendix~\ref{sec:human_evaluation}.

\subsection{Ablation Study}
\label{sec:ablation_study}

Table~\ref{tab:reward_ablation} evaluates the reward branches and auxiliary components. Compared with pixel-only RL, independently adding
the ID VJEPA encoder reward or OOD VJEPA predictor reward improves overall performance, while their combination performs best. Ablations of $\lambda_{\text{VJEPA}}$ and $\alpha_{\text{OOD}}$ are provided in Appendix~\ref{sec:ablation_of_VJEPA}.

Removing static regularization, embodiment-focused rewards, or the
visual fidelity reward consistently degrades performance, confirming
their complementary contributions. Further analyses of
$\lambda_{\text{static}}$, $\lambda_{\text{emb}}$, and
$\lambda_{\text{VF}}$ are reported in
Appendix~\ref{sec:reward_function_coefficient_design}.



\begin{figure*}[!t]
\centering
\includegraphics[width=0.84\textwidth]{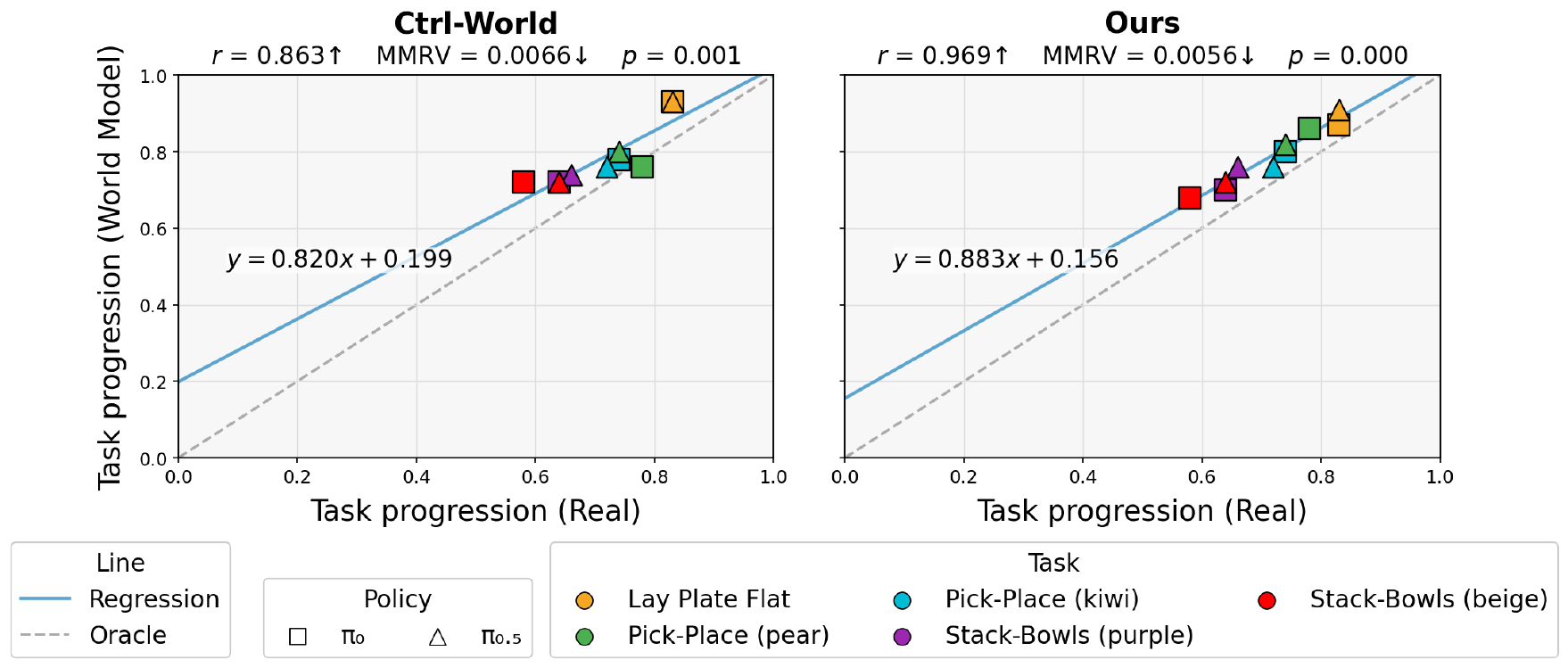}
\caption{Quantitative correlations between real-world and world-model rollouts. 
Our policy predicts 32-action chunks at 30 fps, which are downsampled or extended to match each world model training setting. We evaluate 20 interactions between the policy and the world model.} 
\label{fig:quan_vla_evaluation}
\end{figure*}

\begin{figure*}[!t]
\centering
\includegraphics[width=0.85\textwidth]{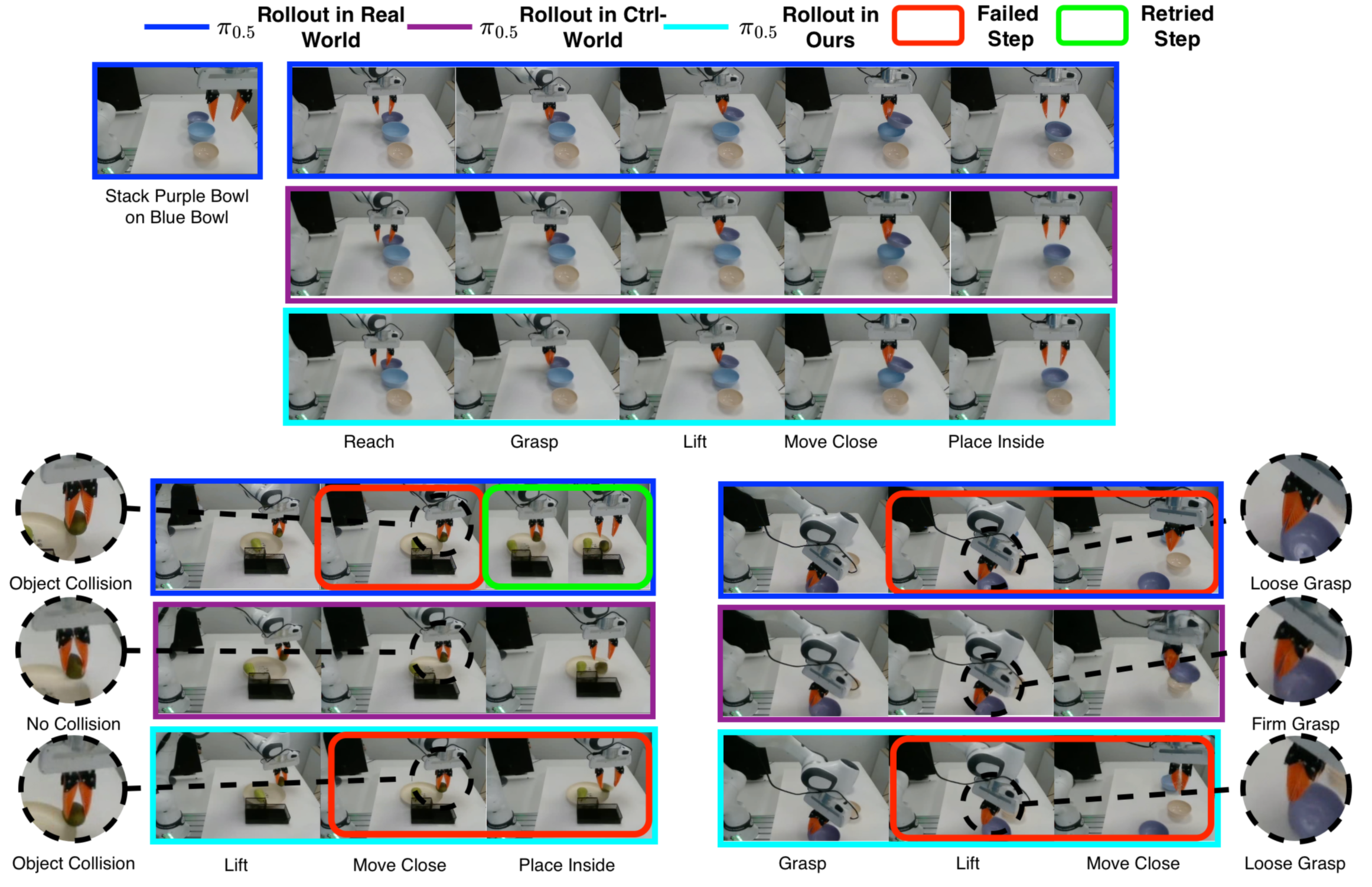}
\caption{Qualitative comparison of $\pi_{0.5}$ rollouts in the real world and world model. 
} 
\label{fig:qual_vla_evaluation}
\end{figure*}

\subsection{VLA Evaluation}

To evaluate our world model as a policy evaluator, we test two Droid-focused policies—$\pi_0$, and $\pi_{0.5}$. We compare our model against Ctrl-world across five tasks: (1) lay plate flat, (2) pick-place (pear), (3) pick-place (kiwi), (4) stack-bowls (purple) and (5) stack-bowls (beige). Each task comprises around expert 300 trajectories to posttrain the policy, the world model and VJEPA-2. Starting from our Droid-pretrained checkpoint, we perform 5k steps of SFT followed by 1k steps of RL, while Ctrl-world is fine-tuned for 7k steps for fairness. 

After training, we evaluate each task over 10 episodes in both the real world and the learned world model. The overall progress is divided into $N$ steps, where successfully completing each step in sequence contributes $1/N$ to the total progress. The lay plate flat task is decomposed into three steps: (1) reach, (2) touch, and (3) flatten. The pick-place task consists of five steps: (1) reach, (2) grasp, (3) lift, (4) move close, and (5) place inside. The stack-bowls task follows the same five-step structure, but differs in the measurement of grasp, and placement-in-bowl phases due to the distinct geometry of bowls compared to fruits.

We report a quantitative comparison of policy rollouts between real world and the world model in Figure~\ref{fig:quan_vla_evaluation}. Results show that, compared to Ctrl-World, our model exhibits reduced overestimation of success rates and provides more accurate estimates of real-world performance. Qualitative results are shown in Figure~\ref{fig:qual_vla_evaluation}, where Ctrl-world often predicts task success even when the policy fails from the same initial observation in real world. While our model faithfully reflects real-world dynamics (e.g., object collisions and loose grasps), these results demonstrate robustness in handling suboptimal policy failures under RL with OOD actions.

\section{Conclusion}

We present \textbf{PAVXploreRL}, a model-agnostic reinforcement learning (RL) pipeline featuring a unified reward function that jointly optimizes the PAV targets: Physical Plausibility, Action Adherence, and Visual Fidelity. The pipeline trains on both in-distribution and out-of-distribution actions without requiring paired video supervision. Empirical results on our pretrained action-conditioned world model demonstrate consistent gains in both quantitative metrics and qualitative evaluations, as well as improved its capability as a reliable action policy evaluator.

A limitation of this work is that our focus is primarily on RL design, while the base world model does not explicitly incorporate components such as multi-view consistency. Another limitation is the long training time; despite using DiffusionNFT as a more efficient RL pipeline, the long rollouts and reward computation still incur substantial computational cost, making it suitable primarily as a post-training method. Nevertheless, the proposed RL framework is orthogonal to these design choices and can be readily extended to future world model architectures. Future work could extend our method to advanced pretrained world models to further test its generality and robustness.



\section*{AI Use Statement}
Generative AI tools were used to support idea discussion, language polishing, and limited code execution and debugging. All research decisions, implementations, experiments, analyses, and conclusions were performed or verified by the authors, who take full responsibility for the content of this paper.

\section*{Reproducibility Statement}
We provide the implementation as supplementary material. The accompanying README includes instructions for training and inference, while additional implementation details and experimental settings are documented in the paper and appendix. Upon acceptance, we will publicly release the trained model weights.

\section*{Ethics Statement}
This work focuses on action-conditioned video generation for robotics research. Our implementation builds upon publicly available code, pretrained models, and research datasets, subject to their respective licenses and usage terms. We conduct a blinded human evaluation but do not collect personal or sensitive participant data. Although we do not anticipate direct harmful impacts, generated videos should be treated as simulated predictions rather than guaranteed representations of real-world outcomes, particularly in safety-critical robotics applications.

\bibliography{iclr2027_conference}
\bibliographystyle{iclr2027_conference}

\newpage
\appendix
\section{Appendix}

\subsection{Visual Fidelity Reward Components}
To evaluate visual fidelity without ground-truth video references, we employ a set of non-reference metrics summarized in Table~\ref{tab:vq_effect}. The overall reward $r_{r_{VF}}$ combines multiple complementary objectives that jointly regularize spatial sharpness, perceptual realism, and temporal consistency.

\textbf{Laplacian} sharpness encourages high-frequency detail by penalizing low variance in the Laplacian response of grayscale frames, thereby discouraging blurriness. \textbf{NIQE}-inspired naturalness measures deviations of MSCN statistics from Gaussianity, capturing perceptual artifacts without reference supervision. \textbf{Flow Distribution} regularizes the statistics of frame-wise intensity changes, penalizing static collapse, excessive dynamics, and degenerate motion patterns to promote physically plausible temporal evolution.

For temporal consistency, \textbf{Temporal Gradient} penalizes second-order temporal derivatives to suppress flickering, while \textbf{Interframe SSIM} enforces structural coherence between adjacent frames via $1-\text{SSIM}$. In the frequency domain, \textbf{Phase Coherence} constrains temporal drift of Fourier phase representations, preserving stable spatial structure over time.

Appearance stability is further enforced via \textbf{Color Stability}, which penalizes temporal shifts in histogram-based color distributions. Finally, \textbf{Gradient Consistency} minimizes temporal variance in Sobel gradient magnitudes, reducing texture flickering and preserving edge stability across frames.

\label{sec:visual_quality_reward}
\begin{table}[ht]
\centering
\small
\caption{Summary of Visual Fidelity reward components and their effects.}
\begin{tabular}{ll}
\toprule
\textbf{Metric} & \textbf{Effect} \\
\midrule
 Laplacian            & Captures frame sharpness \\
 NIQE                 & Measures perceptual quality (no reference) \\
Flow Distribution    & Ensures optical flow consistency \\
Temporal Gradient    & Detects sudden motion changes \\
Interframe SSIM      & Evaluates frame-to-frame similarity \\
 Phase Coherence      & Captures frequency-domain motion consistency \\
Color Stability      & Maintains stable colors \\
Gradient Consistency & Preserves spatial edge consistency over time \\
\bottomrule
\end{tabular}
\label{tab:vq_effect}
\end{table}

\subsection{VJEPA-2 Training Pipeline}
\begin{figure}[H]
\centering
\small
\includegraphics[width=0.88\columnwidth]{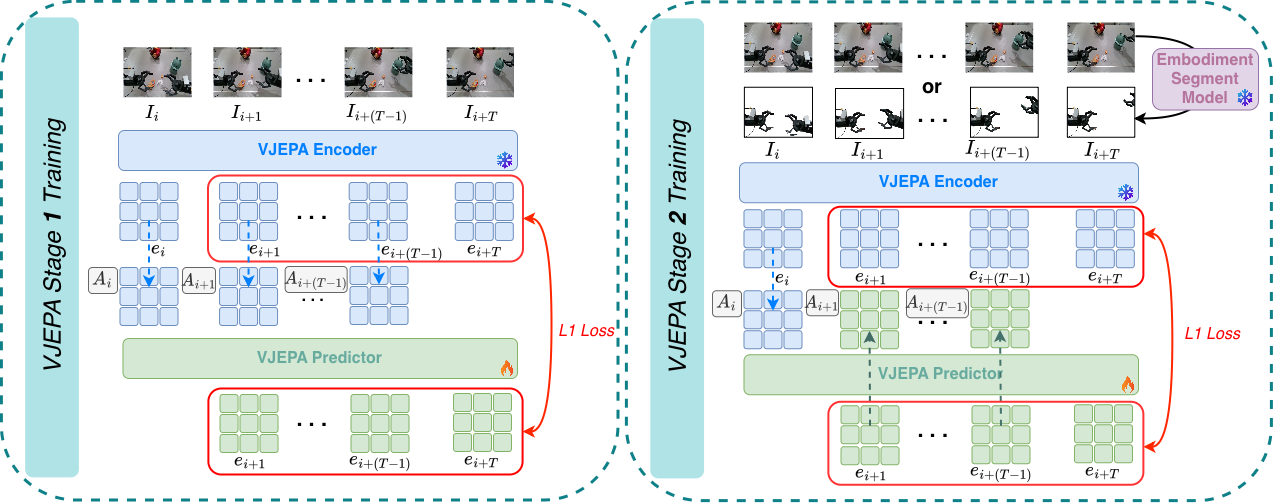}
\caption{VJEPA-2 training pipeline. \textbf{Stage 1:} VJEPA-2 is trained on raw frames with one-step prediction. \textbf{Stage 2:} It extends to autoregressive multi-step prediction over a horizon $T$, iteratively propagating from an initial frame using frames sampled from raw or embodiment-focused views.}
\label{fig:vjepa2_pipeline}
\end{figure}

This is a supplementary illustration of the VJEPA training pipeline shown in Figure~\ref{fig:vjepa2_pipeline}. The full pipeline is described in Section~\ref{sec:vjepa_training}.

\label{sec:vjepa_training_appendix}

\subsection{Model Performance over Training Steps}
\label{sec:model_performance}
\begin{figure}[!t]
\centering
\small

\begin{minipage}{0.48\columnwidth}
    \centering
    \includegraphics[width=\linewidth]{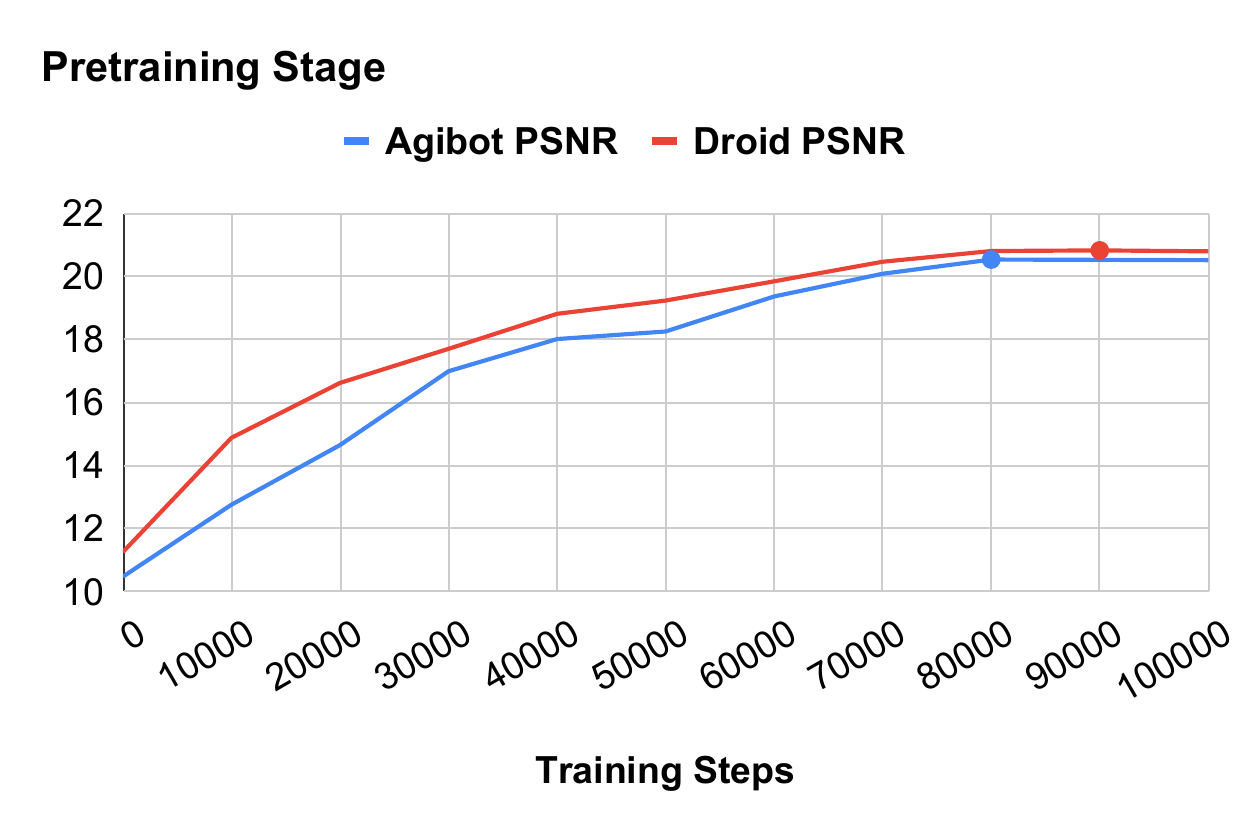}
\end{minipage}
\hfill
\begin{minipage}{0.48\columnwidth}
    \centering
    \includegraphics[width=\linewidth]{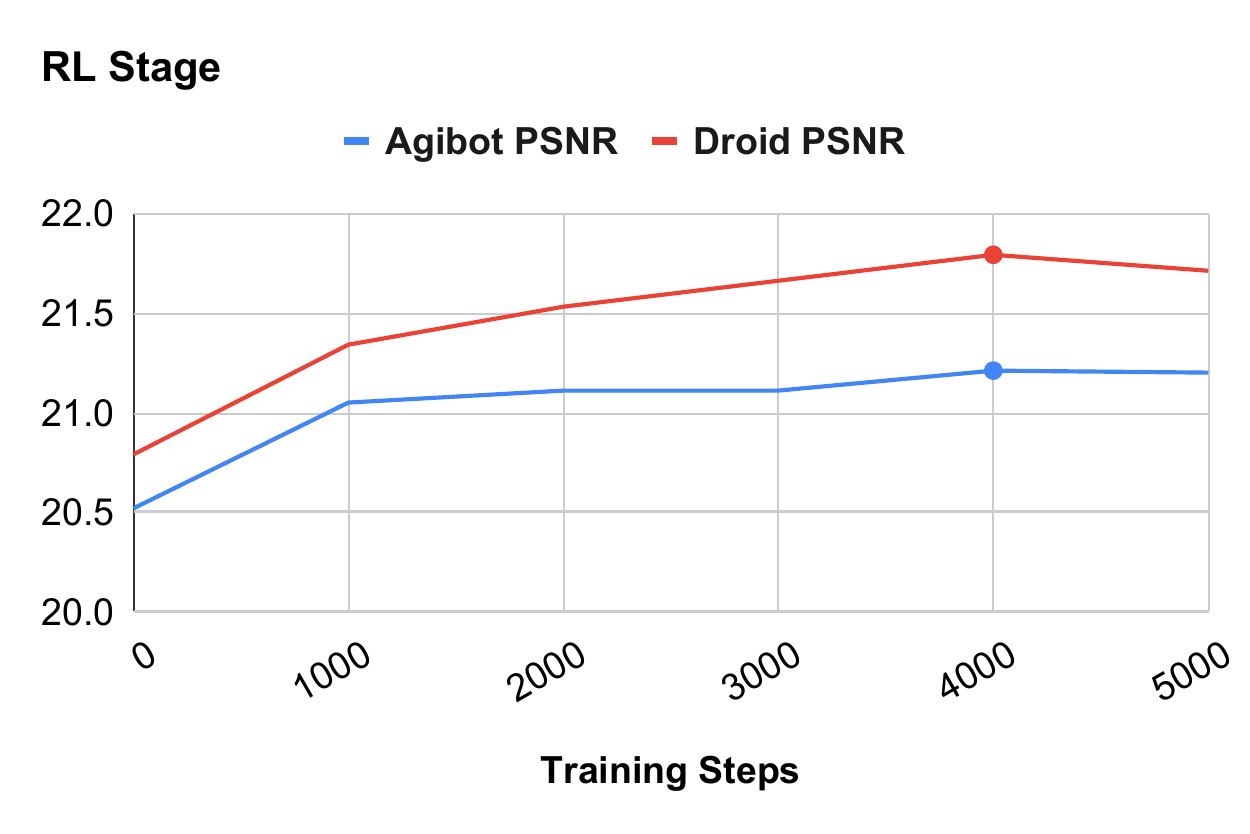}
\end{minipage}

\caption{PSNR of model on validation sets in pretraining and reinforcement learning stage over training steps for Agibot and Droid datasets.}
\label{fig:model_performance}
\end{figure}

We report PSNR on the validation set across training steps for both datasets during the pretraining and reinforcement learning (RL) stages in Figure~\ref{fig:model_performance}. We firstly observe that pretraining converges at approximately 80K steps, with further Supervised Fine-Tuning (SFT) yielding marginal improvements, indicating that the RL posttraining gains do not stem from additional training alone.
 Based on these results, we select 80K steps for pretraining and 5K steps for RL, achieving a balance between performance and efficiency.

\subsection{Human Evaluation}
\label{sec:human_evaluation}

\begin{table*}[!t]
\centering
\small
\setlength{\tabcolsep}{4pt}
\renewcommand{\arraystretch}{1.2}
\caption{
Blinded human evaluation of the pretrained model (PT) and
PAVXploreRL (RL) on 100 randomly sampled AgiBot validation
trajectories. Each trajectory is evaluated by two independent
evaluators. We report trajectory-level average preference rates.
}
\label{tab:human_evaluation}
\begin{tabular}{lcccc}
\toprule
\textbf{Criterion}
& \textbf{RL Preferred $\uparrow$}
& \textbf{Tie}
& \textbf{PT Preferred $\downarrow$}
& \textbf{Cohen's $\kappa$ $\uparrow$} \\
\midrule
Physical Plausibility & \textbf{60.5\%} & 17.0\% & 22.5\% & 0.36 \\
Action Adherence      & \textbf{65.0\%} & 16.5\% & 18.5\% & 0.49 \\
Visual Fidelity       & \textbf{68.0\%} & 14.0\% & 18.0\% & 0.53 \\
Overall Quality       & \textbf{69.0\%} & 16.0\% & 15.0\% & 0.52 \\
\bottomrule
\end{tabular}
\end{table*}

We conduct a blinded human evaluation on 100 trajectories randomly sampled across all 36 tasks in the AgiBot validation set. For each trajectory, two independent evaluators compare PT and RL predictions, presented in randomized order alongside the ground-truth future. They assess physical plausibility, action adherence, visual fidelity, and overall quality, selecting PT, RL, or a tie for each criterion. We report trajectory-level average preference rates and Cohen's $\kappa$ for inter-rater agreement. As shown in Table~\ref{tab:human_evaluation}, the results provide an evaluation of PAV quality independent of the VJEPA-based metrics.

\subsection{Ablation Study of VJEPA Reward}
\label{sec:ablation_of_VJEPA}
As discussed in Section~\ref{sec:ablation_study}, the contribution of VJEPA-E is governed by $\lambda_{\text{VJEPA}}$, while VJEPA-P, due to its critical role in the out-of-distribution (OOD) branch, is effectively controlled by the OOD branch proportion $\alpha_{\text{OOD}}$. We conduct a detailed ablation over these coefficients in Table~\ref{tab:ablation_ratio_full}. We observe that moderate values yield optimal performance, whereas excessively large values can degrade performance. The best performance is achieved with $\lambda_{\text{VJEPA}}=0.2$ and $\alpha_{\text{OOD}}=0.1$.

\begin{table*}[!t]
\centering
\small
\setlength{\tabcolsep}{4pt}
\renewcommand{\arraystretch}{1.2}
\caption{Ablation study on the effect of VJEPA based metrics ratios in reward function, evaluated on Agibot with 1K-step reinforcement learning.}
\begin{tabular}{c|ccc|cc|cc}
\hline
\multirow{2}{*}{Setting}
& \multicolumn{3}{c|}{Visual Quality}
& \multicolumn{2}{c|}{Flow Consistency}
& \multicolumn{2}{c}{VJEPA Loss} \\
\cline{2-8}

& PSNR $\uparrow$ & SSIM $\uparrow$ & LPIPS $\downarrow$
& EPE $\downarrow$ & COS $\uparrow$
& Enc $\downarrow$ & Pred $\downarrow$ \\
\hline

$\lambda_{\text{{VJEPA}}}= 0, \alpha_{\text{{OOD}}} = 0$
& 20.80 & 0.806 & 0.145 & 0.333 & 0.204 & 0.526  & 0.374  \\

\cline{1-8}
\multicolumn{8}{c}{\textit{VJEPA Predictor Only}} \\
\cline{1-8}

$\alpha_{\text{{OOD}}} = 0.05$
& 20.84 & 0.805 & \textbf{0.131} & \textbf{0.325} & \textbf{0.227} & \textbf{0.523} & 0.375 \\

$\alpha_{\text{{OOD}}} = 0.1$
& \textbf{20.94} & \textbf{0.807} & 0.139 & 0.329 & 0.206 & 0.526 & 0.375 \\

$\alpha_{\text{{OOD}}} = 0.2$
& 20.90 & 0.806 & 0.141 & 0.329 & 0.205 & 0.526 & \textbf{0.374} \\

\cline{1-8}

\multicolumn{8}{c}{\textit{VJEPA Encoder Only}} \\
\cline{1-8}

$\lambda_{\text{{VJEPA}}} = 0.1$
& 20.86 & 0.806 & 0.141 & 0.333 & 0.204 & 0.526 & 0.375 \\

$\lambda_{\text{{VJEPA}}} = 0.2$
& \textbf{20.93} & \textbf{0.807} & \textbf{0.135} & \textbf{0.324} & \textbf{0.216} & \textbf{0.524} & \textbf{0.374} \\

$\lambda_{\text{{VJEPA}}} = 0.3$
& 20.82 & 0.805 & \textbf{0.135} & 0.325 & 0.215 & \textbf{0.524} & \textbf{0.374} \\

\cline{1-8}
\multicolumn{8}{c}{\textit{VJEPA Encoder and Predictor}} \\
\cline{1-8}

$\lambda_{\text{{VJEPA}}} = 0.2, \alpha_{\text{{OOD}}} = 0.05$
& 20.99 & 0.806 & 0.138 & 0.327 & 0.210 & 0.525 & 0.375  \\

$\lambda_{\text{{VJEPA}}} = 0.2, \alpha_{\text{{OOD}}} = 0.1$
& \textbf{21.05} & \textbf{0.810} & \textbf{0.126} & \textbf{0.310} & \textbf{0.230} & \textbf{0.521} & \textbf{0.373}   \\

\hline

\end{tabular}
\label{tab:ablation_ratio_full}
\end{table*}

\subsection{Reward Function Coefficient Selection}
\label{sec:reward_function_coefficient_design}

\begin{table}[!t]
\centering
\small
\caption{Ablation of reward function coefficients on the AgiBot validation set, reporting the accuracy of correctly assigning the highest reward to the ground-truth (GT) video among 12 evaluated video variants.}
\begin{subtable}{\linewidth}
\centering
\caption{$\lambda_{\mathrm{static}}$}
\begin{tabular}{c|cccccc}
\toprule
Value & 0.0 & 0.2 & 0.4 & 0.5 & 0.8 & 1.0 \\
\midrule
GT ratio & 0\% & 11\% & 39\% & 67\% & \textbf{82\%} & 67\% \\
\bottomrule
\end{tabular}
\end{subtable}

\vspace{0.6em}

\begin{subtable}{\linewidth}
\centering
\caption{$\lambda_{\mathrm{emb}}$}
\begin{tabular}{c|ccccc}
\toprule
Value & 0.00 & 0.25 & 0.50 & 0.75 & 1.00 \\
\midrule
GT ratio & 79\% & 81\% & \textbf{82\%} & 77\% & 75\% \\
\bottomrule
\end{tabular}
\end{subtable}

\vspace{0.6em}

\begin{subtable}{\linewidth}
\centering
\caption{$\lambda_{\mathrm{VF}}$}
\begin{tabular}{c|ccc}
\toprule
Value & 0.0 & 0.1 & 0.2 \\
\midrule
GT ratio & 75\% & \textbf{82\%} & 71\% \\
\bottomrule
\end{tabular}
\end{subtable}
\label{tab:coefficent_selection}
\end{table}

To avoid the high computational cost of RL training, we first perform reward-function coefficient selection through an offline diagnostic benchmark. The underlying assumption is that a well-designed reward function should consistently assign the highest reward to high-quality videos while penalizing degraded ones.

Specifically, we construct 12 video variants, including: (1) ground-truth (GT) videos, (2) videos predicted by the pretrained world model, (3) noisy videos with global or motion-specific perturbations, (4) temporally truncated videos with the first 0.2, 0.4, or 0.8 seconds removed, (5) temporally slowed videos (2× slower) with different slowdown onset points corresponding to 100\%, 75\%, 50\%, and 25\% of the original trajectory, (6) temporally static videos repeating the first frame.

For each coefficient setting, we treat the reward function as a classifier and report accuracy as the proportion of cases in which the ground-truth (GT) video attains the highest reward among 12 variants. This metric is used to ablate the coefficients $\lambda_{\text{static}}$, $\lambda_{\text{emb}}$, and $\lambda_{\text{VF}}$, with higher accuracy indicating stronger alignment between the reward function and video quality. 

The results are reported in Table~\ref{tab:coefficent_selection}. We observe that $\lambda_{static}$ has the most significant impact and is essential; removing it leads to cases where ground-truth videos consistently receive lower rewards than temporally static videos. Both $\lambda_{emb}$ and $\lambda_{VF}$ further improve performance, indicating that embodiment consistency and visual fidelity act as complementary signals to the main reward, and that moderate weighting improves classifier accuracy. Based on these results, we set $\lambda_{static} = 0.8$, $\lambda_{emb} = 0.5$, and $\lambda_{VF} = 0.1$.

\end{document}